\documentclass[11pt]{article}

\usepackage[preprint]{acl}

\usepackage{times}
\usepackage{latexsym}

\usepackage[T1]{fontenc}

\usepackage[utf8]{inputenc}

\usepackage{microtype}

\usepackage{inconsolata}

\usepackage{graphicx}
\usepackage{booktabs}
\usepackage[table,xcdraw]{xcolor}
\usepackage{colortbl}
\usepackage{multirow}

\usepackage{amsmath}
\usepackage{amssymb}

\usepackage{enumitem}

\title{PretrainRL: Alleviating Factuality Hallucination of Large Language Models at the Beginning}

\author{
\textbf{Langming Liu\textsuperscript{1}}\footnotemark[1],
\textbf{Kangtao Lv\textsuperscript{1,2}}\footnotemark[1], 
\textbf{Haibin Chen\textsuperscript{1}},
\textbf{Weidong Zhang\textsuperscript{1}},
\textbf{Yejing Wang\textsuperscript{3}},\\
\textbf{Shilei Liu\textsuperscript{1}},
\textbf{Xin Tong\textsuperscript{1}},
\textbf{Yujin Yuan\textsuperscript{1}},
\textbf{Yongwei Wang\textsuperscript{2,4}},
\textbf{Wenbo Su\textsuperscript{1}},
\textbf{Bo Zheng\textsuperscript{1}}\footnotemark[2]\\
\textsuperscript{1}Future Living Lab of Alibaba 
\textsuperscript{2}Zhejiang University \\
\textsuperscript{3}Alibaba Group 
\textsuperscript{4}Shanghai AI Laboratory \\
\texttt{\{liulangming.llm, lvkangtao.lkt, bozheng\}@alibaba-inc.com} \\
}

\begin{document}
\begingroup
\renewcommand{\thefootnote}{\fnsymbol{footnote}}
\maketitle
\footnotetext[1]{Equal contribution}
\footnotetext[2]{Corresponding author}
\endgroup
\begin{abstract}
Large language models (LLMs), despite their powerful capabilities, suffer from factual hallucinations where they generate verifiable falsehoods. We identify a root of this issue: the imbalanced data distribution in the pretraining corpus, which leads to a state of "low-probability truth" and "high-probability falsehood". Recent approaches, such as teaching models to say "I don't know" or post-hoc knowledge editing, either evade the problem or face catastrophic forgetting. To address this issue from its root, we propose \textbf{PretrainRL}, a novel framework that integrates reinforcement learning into the pretraining phase to consolidate factual knowledge. The core principle of PretrainRL is "\textbf{debiasing then learning}." It actively reshapes the model's probability distribution by down-weighting high-probability falsehoods, thereby making "room" for low-probability truths to be learned effectively. To enable this, we design an efficient negative sampling strategy to discover these high-probability falsehoods and introduce novel metrics to evaluate the model's probabilistic state concerning factual knowledge. Extensive experiments on three public benchmarks demonstrate that PretrainRL significantly alleviates factual hallucinations and outperforms state-of-the-art methods.

\end{abstract}

\section{Introduction}
LLMs \cite{openai2024gpt4technicalreport,deepseekai2024deepseekv3technicalreport,grattafiori2024llama3herdmodels,yang2025qwen3technicalreport} have recently drawn significant attention for their strong capabilities in natural language processing (NLP). Although LLMs show powerful knowledge retention, LLMs suffer from ubiquitous and difficult-to-cure hallucination issues. The taxonomy of such hallucinations is constantly evolving, generally categorized as factuality-related, instruction-related, and logic-related errors~\cite{ji2023survey,tonmoy2024comprehensive,zhang2025siren}. In this work, we focus on \textbf{factual hallucinations}, where model generations contain clear and verifiable errors. This type of hallucination is arguably the most fundamental and has the most extensive impact, since it pertains to statements that can be easily disproven with ground-truth knowledge. For example, when the user ask ``What is the capital of Arno?'', and LLM response is that ``Rome'', however the answer is ``Florence''.

\begin{figure}[t]
\begin{center}
\centerline{\includegraphics[width=\columnwidth]{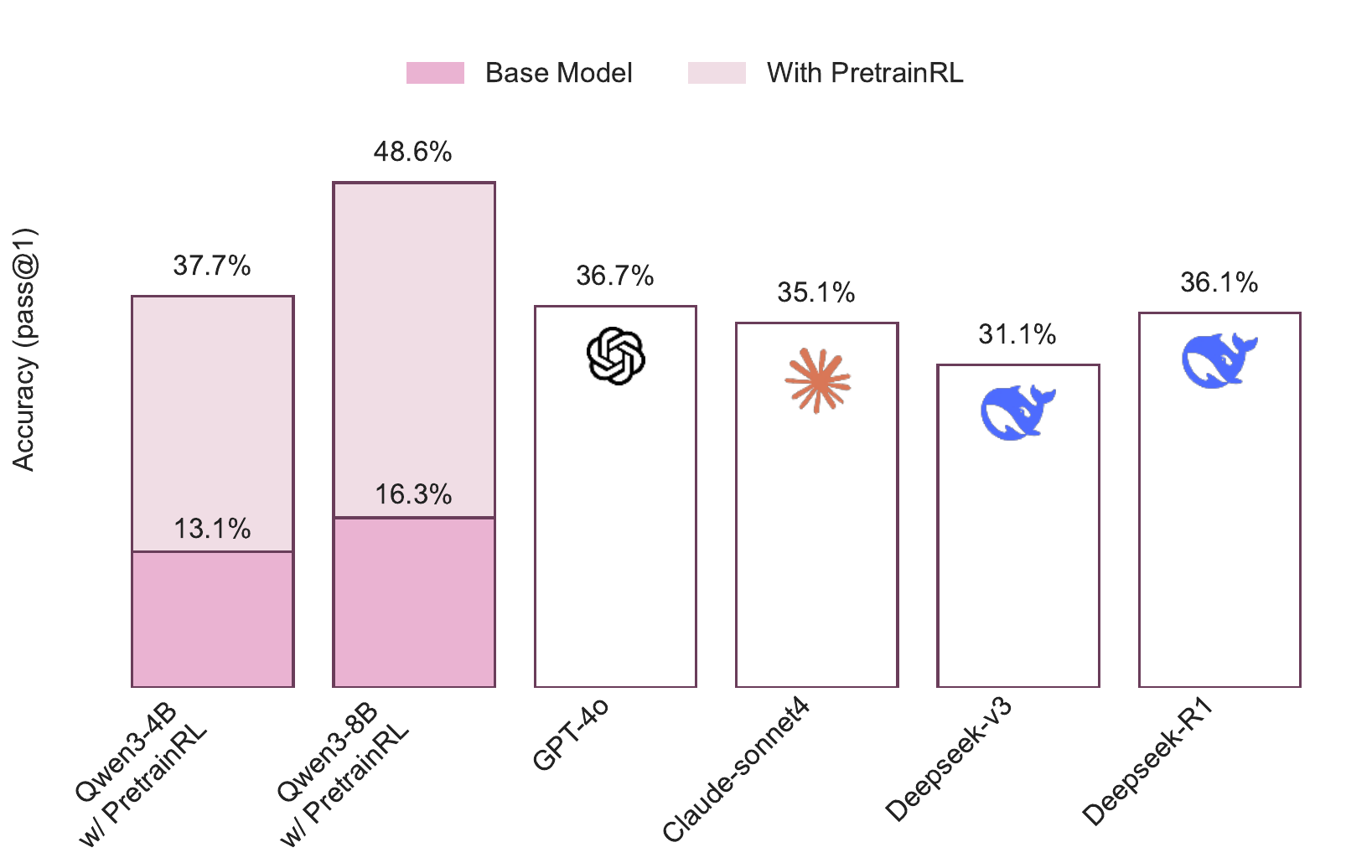}}
\vskip -0.1in
\caption{Overall comparison on POPQA}
\label{fig:POPQA_performance_compare}
\end{center}
\vskip -0.4in
\end{figure}

Several efforts have been made to address or alleviate factual hallucinations. A prevalent approach is to teach LLMs to say ``\textbf{I don't know}'' (IDK) when uncertain \cite{lin2022teaching,cao2023learn,manakul2023selfcheckgpt}. This involves learning a preference for refusing to answer when generation probabilities are low, implemented through preference alignment, novel decoding methods, or specialized loss functions. While this direction is safe and effective at avoiding hazardous responses, it has notable drawbacks: models can become ``lazy'' by over-producing IDK responses, which in turn restricts generative diversity \cite{kadavath2022language,touvron2023llama,bai2022training}. Crucially, this method cannot directly resolve factual hallucinations, but merely conceals them. Another line of research, known as \textbf{knowledge editing} \cite{decao2021editingfactualknowledgelanguage,meng2022locating,meng2022mass,zheng2023can,wang2024knowledge}, aims to efficiently and precisely correct factual errors. This involves locating and making subtle modifications to the parameters associated with a specific fact. However, the core challenge lies in its scalability; a large number of edits can degrade overall capabilities and lead to catastrophic forgetting~\cite{mitchell2021fast}.

Although teaching models to say IDK does not address factual hallucinations at its root, it points to a core source: truths have a low generation probability. More accurately, \textbf{low-probability truth} and \textbf{high-probability falsehood} jointly cause hallucinations. This phenomenon exists widely in LLMs, and the in-depth reason lies not in noisy or polluted data, but in the pretraining paradigm itself \cite{xu2024hallucination,kalai2025languagemodelshallucinate}. The next-token prediction (NTP) objective forces the model to fit the corpus data distribution, which is essentially a density estimation~\cite{radford2018improving}. However, an imbalanced data distribution is inevitable \cite{gao2020pile,roberts2020much}. For certain knowledge domains, this imbalance is extreme: a small amount of ``head knowledge'' occupies a high cumulative density. In this scenario, head knowledge can interfere with the memorization of ``tail knowledge'' during training, causing ``head falsehoods'' (i.e., high-probability falsehoods). For instance, if a corpus contains factual knowledge of ``Brand A shoes are red'' 1,000 times more frequently than ``Brand B shoes are blue,'' an LLM may learn a shortcut to associate ``shoes are'' with ``red,'' degrading from learning a conditional probability to a marginal one and leading to severe hallucinations. This kind of degradation can also occur for more common data distributions.

The critical challenge is to consolidate this tail knowledge without disrupting existing knowledge. Modifying the corpus distribution is ill-advised as it can introduce new biases. Post-training is also inappropriate, as the model's knowledge is more ``rigid'' at this stage, and aggressive knowledge injection can cause catastrophic forgetting or pattern collapse. Therefore, we propose \textbf{Pretrain}ing \textbf{R}einforcement \textbf{L}earning, a framework that consolidates knowledge during the pretraining stage. The core principle of PretrainRL is ``debiasing then learning'': first lowering the probability of falsehoods to ``make room,'' then increasing the probability of truths. Unlike simply continual pretraining~\cite{wang2024comprehensive,shi2024continual,ke2023continual}, which focuses only on learning, this debiasing step is crucial given the limited capacity of parametric knowledge storage. We leverage direct preference optimization (DPO) \cite{rafailov2023direct} for this ``probability reshaping.'' The main challenge then becomes discovering the rejected samples (i.e., head falsehoods) for DPO, which determines the room for truth learning. We design an efficient sampling approach to obtain high-quality rejected samples for DPO. Furthermore, we introduce three metrics to better monitor and evaluate the state of the model's generation probability. The contributions of our work are three-fold as follows:

\begin{itemize}[leftmargin=*]
\item We reveal that the core source of factual hallucination is the imbalanced distribution in the pretraining corpus and propose PretrainRL, which novelly introduces an RL method to alleviate factual hallucinations during the pretraining phase.
\item We propose an efficient and effective method for negative sampling, boosting DPO's effectiveness on probability reshaping. We also introduce crucial metrics for a comprehensive evaluation of the model's probabilistic state.
\item Thorough and extensive experiments on three public benchmarks demonstrate the superior performance of the proposed PretrainRL compared to other state-of-the-art LLMs. Further analysis provides insights into the study of hallucination.
\end{itemize}

\section{Preliminary}

\subsection{Knowledge Graph}
A knowledge graph (KG) is a structured representation of factual knowledge, where entities (nodes) are interconnected by relations (edges) \cite{hogan2021knowledge,chen2020knowledge}. 
KGs serve as a foundational resource for a wide range of NLP tasks by providing explicit, verifiable world knowledge.
The basic unit of a KG is the \textbf{knowledge triple}, a structured assertion of a fact in the form of $(s, p, o)$.
Here, $s$ is the \textbf{subject} (i.e., head entity), $o$ is the \textbf{object} (i.e., tail entity), and $r$ is the \textbf{predicate} (i.e., relation that connects them).
For example, the factual statement ``Paris is the capital of France'' can be represented as the triple \texttt{(Paris, capitalOf, France)}. 
The subject \texttt{Paris} and object \texttt{France} are entities, while \texttt{capitalOf} is the relation describing their semantic connection.
This triple-based structure allows for the systematic organization of complex information.
In the context of LLMs, KGs offer a crucial, external source of ground-truth information, providing a pathway to verify model-generated statements and mitigate factual hallucinations.

\subsection{Next Token Prediction}

The predominant pretraining objective for auto-regressive LLMs is \textbf{Next-Token Prediction} (NTP).
Given a sequence of tokens $x = \{x_1, x_2, \dots, x_{i-1}\}$, the model is trained to predict the next token $x_i$ by maximizing its conditional probability. 
Formally, the objective is to minimize the negative log-likelihood (NLL) over a large-scale pretraining corpus $\mathcal{D}$:
\begin{equation}
\mathcal{L}_{\text{NTP}} = - \mathbb{E}_{x \sim \mathcal{D}} \sum_{i=1}^{|x|} \log P(x_i | x_{<i}; \theta)
\end{equation}
where $x$ is a sequence in the corpus, $x_{<i}$ denotes the preceding tokens, and $\theta$ represents the model parameters.
This self-supervised objective compels the base model to internalize a large amount of syntactic, semantic, and factual knowledge from the training data in order to make accurate predictions.
By fitting the data distribution of the pretraining corpus (i.e., density estimation), the model essentially learns a statistical representation of the world.
However, the model's learned distribution is susceptible to biases present in the data, such as the frequency imbalance of factualities, which is a primary source of hallucination.

\section{Methodology}

We first analyze the source of the factual hallucination of LLMs, then provide the solution to address this issue.

\subsection{Source of Factual Hallucination}
The KG triple can be considered to be the basic unit of the factuality in corpus.
Suppose that the triple $(s,p,o)$ is a factual knowledge in sequence $x$. Then, minimizing NLL loss can also maximize 
\begin{equation}
    P(o | s, p; \theta), 
\end{equation}
which is a conditional probability of the object.
As in the previous statement, the critical source of factual hallucinations is low-probability truth and high-probability falsehood. To put it more deeply, the imbalanced distribution of knowledge causes such probability error of LLMs.

Given a certain knowledge category, suppose that this imbalanced data structure consists of only two types of triples, $D_1=(s_1, p, o_1)$, $D_2=( s_2, p, o_2)$. The frequency of $D_1$ far exceeds that of $D_2$, which means that frequency $\xi_1$ may be hundreds of frequency $\xi_2$ in the corpus. We call $D_1, D_2$ \textbf{head knowledge} and \textbf{tail knowledge}, respectively. The NLL loss is then partially maximizing the weighted sum of conditional probability of head and tail knowledge, formed as
\begin{equation}
    \xi_1 P(o_1 | s_1, p; \theta)+ \xi_2 P(o_2 | s_2, p; \theta), 
\end{equation}
where $\xi_1,\xi_2$ is the frequency, and $\xi_1\gg\xi_2$. As such an imbalanced knowledge exists widely in corpus, the pretraining task (i.e., NTP) may force LLMs to take shortcuts to obtain this objective. Specifically, LLMs tend to neglect the subject $s$ of the context, instead degrading to learn a margin $P(o | p; \theta)$ of conditional probability. In this case, LLM have a high probability $P(o_1 | p; \theta)$ to predict object $o_1$, regardless of given context $s_1$ or $s_2$. Consequently, head knowledge $D_1$ successfully permeates tail knowledge $D_2$, leading to severe factual hallucination.
From the perspective of the attention mechanism, LLMs learn high attention scores on predicate $p$, and low attention scores on subject $s$. This is the reason that, even though we emphasize the subject in the prompt, LLMs will persist in outputting a popular answer rather than the correct answer.

\subsection{Negative Sampling}
\label{sec:neg_sample}
The previous theoretical analysis points out that head knowledge $D_1$ squeezes the memory room of tail knowledge $D_2$, resulting in very low conditional probability $P(o_2 | s, p; \theta)$ so as to degenerate to a marginal probability. To address this issue, the primary task is to locate those head knowledge (i.e., negative samples) that affect LLMs' memory of the tail and need to be debiased. The most intuitive way is to classify knowledge into multiple categories, and then use the popularity or frequency of the knowledge in each category to obtain the quantiles of knowledge. The quantiles decide whether the knowledge belongs to head.
However, the pretraining corpus is usually not available for us. Even if we can access the corpus, calculating popularity of knowledge is complicated, since the forms of text are irregular. 

Therefore, we propose an efficient sampling method to obtain negative samples, using the \textbf{beam search} decoding method. Specifically, given a knowledge triple $(s,p,o)$, we use a prompt consisting of $s,p$ and ask LLMs to continue (see Appendix~\ref{sec:Prompt} for prompt details), where we use beam search to let LLM decode several answers. Then we obtain the probability of each answer using the probability of cumulative multiplication of all tokens. Notice that ground truth $o$ should be filtered out. We order the answers with probability and get the negative samples using a threshold or a preset quantity (e.g., $5, 10$). The negative samples obtained directly reflect the LLM's inherent bias for head knowledge, which are actually our targets. 

Furthermore, since models inherently prioritize generating head knowledge for a given question category, a subset of the full dataset suffices to statistically approximate the head knowledge distribution. Taking EntityQuestions (170k instances) as an example, computing statistics over the entire dataset is computationally redundant. We sample 1k instances per \textbf{question category} and count the top 20 high-frequency object, then compare this distribution with the top 20 obtained from the full dataset. The cosine similarity between the two frequency distributions is 0.88, and the Spearman rank correlation is 0.85, indicating that the sampled distribution closely matches the full-data distribution and confirming the effectiveness of our approach.

\subsection{Optimization: PretrainRL}
Our target is to reshape the model's probabilities of knowledge to avoid factual hallucinations. As is well known, RL is designated for lossless probability reshaping. In addition, selecting the specific optimization approach and timing is also crucial.

\subsubsection{Methods: DPO}
We can easily obtain the negative samples (i.e., using sampling methods as mentioned in Section~\ref{sec:neg_sample}) and positive samples (i.e., the ground truth) of knowledge. 
Therefore, the natural approach is to use direct preference optimization (DPO) for probability reshaping. 
As factual knowledge is verifiable, reinforcement learning with verifiable rewards (RLVR) methods \cite{lambert2024tulu,shao2024deepseekmath} is an alternative. However, RLVR is inefficient in the pretraining stage, as the amount of knowledge is gigantic. Moreover, DPO matches more closely with our proposed efficient negative sampling methods in Section~\ref{sec:neg_sample}. Thus, we choose DPO as the optimization method to mitigate factual hallucinations.

\begin{table*}[th]
\small
\setlength{\tabcolsep}{3pt}
\centering
\caption{Overall Comparison}
\vskip -0.1in
\resizebox{ \textwidth}{!}{%
\begin{tabular}{l|cccc|cccc}
\toprule
\multirow{2}{*}{\textbf{Method}} & \multicolumn{4}{c|}{\cellcolor{lightgray}\textbf{\textit{PopQA}}} & \multicolumn{4}{c}{\cellcolor{lightgray}\textbf{\textit{Wikidata-knowledge infusion}}} \\
 & \cellcolor{lightgray}\textbf{ACC (\%)} & \cellcolor{lightgray}\textbf{HR (\%)} & \cellcolor{lightgray}\textbf{MRR (\%)} & \cellcolor{lightgray}\textbf{Prob (\%)} & \cellcolor{lightgray}\textbf{ACC (\%)} & \cellcolor{lightgray}\textbf{HR (\%)} & \cellcolor{lightgray}\textbf{MRR (\%)} & \cellcolor{lightgray}\textbf{Prob (\%)} \\
\midrule
Qwen3-4B & 13.05 & 31.60 & 15.17 & 12.38 & 20.65 & 66.96 & 31.49 & 4.44 \\
Prompt (CoT) & 10.92 & 32.47 & 14.78 & 10.92 & 20.02 & 65.97 & 30.12 & 4.24\\
DPO \textit{initialized from Qwen3-4B}& 9.67 & 18.06& 11.02& 33.88 & 33.84 & 47.46 & 37.54 & 18.42 \\
Continual Training (CT) & 29.97 & 60.36 & 32.94 & 20.92 & 37.42 & 79.32 & 45.89 & 2.57  \\
Iterative RPO & 32.50 & 49.30 & 25.12 & 20.77 & 30.23 & 72.23 & 41.63 & 1.69 \\
PretrainRL(ours) & 37.69 & 66.39 & 40.51 & 31.42 & 46.20 & 78.98 & 54.11 & 7.16 \\
\midrule
Qwen3-8B & 16.32 & 37.06 & 18.60 & 18.60 & 23.96 & 70.43 & 34.81 & 5.51 \\
Prompt (CoT) & 15.54 & 37.57 & 17.97 & 12.14 & 23.96 & 70.43 & 34.81 & 5.51 \\
DPO \textit{initialized from Qwen3-8B}& 13.08& 22.17& 13.92& 42.11& 38.51 & 49.49 & 41.70 & 51.41 \\
Continual Training (CT) & 38.53 & 69.17 & 41.93 & 27.50 & 31.45 & 78.78 & 41.47 & 4.51  \\
Iterative RPO  & 41.23 & 48.13 & 32.28 & 40.83 & 33.50 & 75.13 & 44.71 & 1.98 \\
PretrainRL(ours) & 48.61 & 74.92 & 51.51 & 40.48 & 47.58 & 77.10 & 55.98 & 11.10 \\
\midrule
Llama3-8B & 23.66 & 48.69 & 26.35 & 23.10 & 27.60 & 74.14 & 38.43 & 7.54 \\
Prompt (CoT) & 24.28 & 49.68 & 27.06 & 21.32 & 27.71 & 73.72 & 37.77 & 6.15 \\
DPO \textit{initialized from Llama3-8B}& 15.68 & 26.02 & 16.93 & 42.28 & 54.42 & 62.54 & 57.12 & 28.39 \\
Continual Training (CT) & 13.40 & 33.18 & 16.53 & 20.10 & 53.16 & 89.35 & 63.70 & 18.33 \\
Iterative RPO & 33.83 & 32.63 & 22.64 & 28.17 & 58.92 & 79.72 & 59.24 & 5.34 \\
PretrainRL(ours) & 50.16 & 51.38 & 38.63 & 30.67 & 64.69 & 80.79 & 70.74 & 11.01 \\
\bottomrule
\end{tabular}
}
\label{tab:Overall Comparison}
\vskip -0.1in
\end{table*}

\subsubsection{Timing: Pretraining}
RL or SFT methods are typically applied in the post-training stage. However, as discussed in the introduction, post-training often serves preference learning, rather than large-scale knowledge injection. Since the posttrained model is resistant to knowledge injection (Table~\ref{tab:compare_base_instruct}), decreasing the upper bound. Moreover, intensive knowledge learning in the posttraining phase can lead to catastrophic forgetting, as demonstrated in Section~\ref{sec:general_study}. In contrast, pretrained models contact massive and diverse data, and possess unparalleled breadth and coverage of knowledge. Moreover, pretrained models have better plasticity, making it easier for us to perform large-scale probability reshaping \cite{dong2025reinforcement}. So we select the pretraining stage to perform RL methods for debiasing. Consequently, this approach is named \textbf{PretrainRL}. Specifically, we implement PretrainRL during the continual training (CT) phase.

\subsubsection{Objective}

For a KG triple $(s,p,o)$, our goal is to unlearn negative samples (i.e., debiasing the falsehood) and train positive samples (i.e., learning the truth).
We apply general DPO loss by masking the prompt $x_{s,p}$(i.e., synthesized from $s,p$), focusing on learning the object $o$. The DPO loss can be formed as follows: 
\begin{equation}
\begin{aligned}
\label{eq:dpo_loss}
& \mathcal{L}_{\text{DPO}} = - \mathbb{E}_{(s,p,o)\sim\mathcal{D}} \\
& \log \sigma \left( \beta \log
\frac{\pi_\theta(o_w|x_{s,p})}{\pi_{\text{ref}}(o_w|x_{s,p})} - \beta \log \frac{\pi_\theta(o_l|x_{s,p})}{\pi_{\text{ref}}(o_l|x_{s,p})} \right),
\end{aligned}
\end{equation}
where $\pi_\theta$ is the policy model being optimized, while $\pi_{\text{ref}}$ is a frozen reference model (i.e., a pretrained model). The hyperparameter $\beta$ controls the strength. This objective encourages the model $\pi_\theta$ to increase the relative log-probability of the winner response $o_w$ (i.e., ground truth $o$) compared to the loser response $o_l$ (i.e., negative sample related to $o$).

However, two challenges restrict the effectiveness of DPO in probability reshaping. First, when the conditional probability $P(o_w|s,p;\theta)$ is very low, DPO forces $P(o_l|s,p;\theta)$ to be lower, instead of increasing $P(o_w|s,p;\theta)$. Second, simply applying DPO is not coordinated with the pretraining paradigm, raising the risk of pattern collapse. 
To maintain the model's core language capabilities, meanwhile, enhancing DPO's effects to boost probability reshaping, we incorporate a standard NLL loss~\cite{dubey2024llama} (i.e., CT loss). It is formulated as:
\begin{equation}
\label{eq:sft_loss}
\mathcal{L}_{\text{CT}} = - \mathbb{E}_{(s,p,o) \sim \mathcal{D}} \sum_{t=1}^{|x_{s,p,o}|} \log \pi_\theta(x_i | x_{<i}). 
\end{equation}
This CT loss serves as a regularization term, anchoring the model's distribution to the high-quality chosen responses (i.e., ground truth $o$).
The final loss function is a linearly weighted sum of the two components:
\begin{equation}
\label{eq:combined_loss}
\mathcal{L}_{\text{PretrainRL}} = \mathcal{L}_{\text{DPO}} + \lambda \mathcal{L}_{\text{CT}}
\end{equation}



\subsection{Evaluation metrics}
\label{sec:metrics}
The previous analysis indicates that the imbalanced training of knowledge results in low conditional probability $P(o | s, p; \theta)$ of tail knowledge. To evaluate the mastery of knowledge, we can ask LLM to complete a continuation or QA tasks given context $s,p$. However, traditional metrics, such as accuracy, cannot intuitively evaluate the overall state of probabilities. Therefore, we apply the \textbf{beam search} decoding method, aligning with proposed negative sampling, and introduce three metrics to comprehensively address this challenge. Suppose the beam size is $k$, we define the metrics as follows.
\begin{itemize}[leftmargin=*]
\item \textbf{Hit Ratio (HR@k):} HR is a binary metric that indicates whether LLMs output at least one correct answer among all beam search responses. It demonstrates the diversity, which is commonly high for a base model.
\item \textbf{Mean Reciprocal Rank (MRR@k):} MRR evaluates a beam search list by averaging the reciprocal of the rank (ordered by Prob) of the first correct answer across all responses. This is a metric that balances between Prob and HR.
\item \textbf{Probability (Prob@k)} This value reflects the model's confidence in a particular continuation. It only counts when the responses of beam search hit the correct answer.
\end{itemize}
Equipped with these metrics (i.e., Acc, HR, MRR, Prob), we can comprehensively assess the LLMs' mastery of knowledge. Furthermore, such metrics are also friendly to base model, which is beneficial for exploration of hallucination-mitigation methods in pretraining phase. 

We demonstrate Qwen's mastery of knowledge on POPQA, using the above metrics. The results show that the metric HR is high, indicating that Qwen has trained this knowledge. Nevertheless, the metrics MRR and Prob are relatively low, leading to a low accuracy of this knowledge (i.e., factual hallucination). This phenomenon is much more severe for tail knowledge, according to the previous discussion of tail and head knowledge. Moreover, scaling up the model size has limited effect on mitigating this issue.

\section{Experiments}

\subsection{Experimental Setup}
\paragraph{Datasets}
We use POPQA \cite{mallen-etal-2023-trust}, Wikidata-knowledge infusion \cite{lv2025injectknowledgeefficientlyknowledge}, and EntityQuestions \cite{sciavolino2022simpleentitycentricquestionschallenge} as the datasets in our experiments. These datasets primarily focus on evaluating the factual knowledge of models, and all of them exhibit long-tail distributions. More detailed datasets are provided in Appendix \ref{sec:appendix Dataset Description}.

\paragraph{Models}
We select the Qwen3-4B, Qwen3-7B, Qwen3-14B \cite{yang2025qwen3technicalreport} and Llama3-8B \cite{grattafiori2024llama3herdmodels} for our experiments. In our experiments, we utilize the base versions of all models, which have not undergone instruction tuning or other forms of post-training. As discussed earlier, our method focuses on consolidating factual knowledge during the pretraining stage while minimizing impacts on other knowledge. In contrast, instruct-tuned models have undergone extensive post-training, which make their embedded knowledge more rigid and difficult to modify. Besides, fine-tuning instruct model on such tasks often yields marginal improvements and can sometimes lead to performance degradation \cite{wu2025shadowfttuninginstructmodel}. We apply our method to both base and instruct-tuned models. The results in Table \ref{tab:compare Base and Instrcut} demonstrate that Base-version models provide greater "learnable room" for improvement compared to instruct models when applying our approach.

\begin{table}[th]
\small
\setlength{\tabcolsep}{3pt}
\centering
\caption{Compare Base and Instrcut Model}
\label{tab:compare_base_instruct}
\vskip -0.1in
\resizebox{ \columnwidth}{!}{%
\begin{tabular}{l | c c c c}
\toprule
Method & \textbf{ACC (\%)} & \textbf{HR (\%)} & \textbf{MRR (\%)} & \textbf{Prob (\%)} \\
\midrule
Qwen3-4B-Instruct & 10.70 & 24.09 & 12.32 & 32.30 \\
~~~~ {\em w/ PretrainRL} & 33.21 & 60.62 & 35.93 & 36.90 \\
Qwen3-4B-Base & 13.05 & 31.60 & 15.17 & 12.38 \\
~~~~ {\em w/ PretrainRL} & 37.69 & 66.39 & 40.51 & 31.42 \\
\midrule
Qwen3-4B-Instruct & 18.09 & 55.17 & 26.44 & 20.83 \\
~~~~ {\em w/ PretrainRL} & 41.59 & 73.62 & 50.94 & 24.21 \\
Qwen3-4B-Base & 20.65 & 66.96 & 31.49 & 4.44 \\
~~~~ {\em w/ PretrainRL} & 46.20 & 78.98 & 54.11 & 7.16 \\
\bottomrule
\end{tabular}
}%
\label{tab:compare Base and Instrcut}
\vskip -0.2in
\end{table}

\paragraph{Implementation Details}
In our method, we first evaluate the base model on the POPQA \cite{mallen-etal-2023-trust}, Wikidata-Knowledge Infusion \cite{lv2025injectknowledgeefficientlyknowledge}, and EntityQuestions \cite{sciavolino2022simpleentitycentricquestionschallenge} datasets, while simultaneously recording the model’s own knowledge recall for these questions. Following the procedure described in Section 3, we then compute the high-frequency attribute values the model tends to produce for this class of questions and use them as the pool of negative candidates. In all experiments, for each training instance we randomly sample \textbf{five negatives} from the top 20 candidate pool, yielding five positive–negative pairs for each question.we repeat the POPQA dataset three times during training, expanding it from 13k to 38k samples. Similarly, the EntityQuestions dataset is repeated twice, increasing its size from 17k to 34k samples. All experiments are conducted for one epoch of training. We use a simple template ``\{Question\}\{pos\}<pad>\{Question\}\{neg\}'' to format all of preference pairs (more details can be seen Appendix \ref{sec:Prompt}). We employ the AdamW optimizer with a fixed weight decay of 0.1. The learning rate is set to 3e-5 for POPQA and Wikidata-Knowledge Infusion, and to 2e-5 for EntityQuestions. The coefficient $\beta$ in the DPO loss is tuned in {0.05, 0.1, 0.5, 1.0}, and we end up using 0.1 in this experiment. For a fair comparison, all experiments share identical training hyperparameters. After training, we evaluate model performance using ACC, Hit, MRR, and Prob with beam size $k=50$, as defined in Section~\ref{sec:metrics}. We evaluate all models with the temperature set to 0. All training was conducted on eight H800 GPUs.
\subsection{Main Results}
PretrainRL improves over baselines. We compare against Chain-of-Thought(CoT) \cite{wei2023chainofthoughtpromptingelicitsreasoning}, Direct Preference Optimization(DPO) \cite{rafailov2023direct}, Continued Training(CT), and iterative RPO \cite{pang2024iterative}. We apply the standard DPO to the same set of preference pairs, initializing from the Qwen3-4B-Base. We also evaluate several mainstream closed-source LLMs, which have larger parameter scales and exhibit advanced reasoning capabilities. As shown in Figure \ref{fig:POPQA_performance_compare}, despite their impressive capabilities, these closed-source LLMs perform poorly on long-tail datasets like POPQA, exhibiting high hallucination rates \cite{sun2024headtotailknowledgeablelargelanguage}. In Table \ref{tab:Overall Comparison}, we demonstrate that PretrainRL substantially enhances the model’s factual memorization and significantly outperforms baselines as well as the evaluated closed-source LLMs. Notably, the improvements are even more pronounced for long-tail distribution dataset.

\begin{table}[h]
\small
\setlength{\tabcolsep}{3pt}
\centering
\caption{Scaling model size}
\vskip -0.1in
\resizebox{ \columnwidth}{!}{%
\begin{tabular}{l | c c c c}
\toprule
Method & \textbf{ACC (\%)} & \textbf{HR (\%)} & \textbf{MRR (\%)} & \textbf{Prob (\%)} \\
\midrule
Qwen3-4B & 13.05 & 31.60 & 15.17 & 12.38 \\
~~~~ {\em w/ PretrainRL} & 37.69 & 66.39 & 40.51 & 31.42 \\
\midrule
Qwen3-8B & 16.32 & 37.06 & 18.60 & 18.60 \\
~~~~ {\em w/ PretrainRL} & 48.61 & 74.92 & 51.51 & 40.48 \\
\midrule
Qwen3-14B & 19.41 & 41.57 & 21.90 & 20.10 \\
~~~~ {\em w/ PretrainRL} & 63.44 & 84.28 & 65.58 & 48.08 \\
\bottomrule
\end{tabular}
}%
\label{tab:Scaling model size}
\vskip -0.2in
\end{table}

\subsection{Scalability Study}
To further validate the effectiveness of our method, we conduct experiments on a larger-parameter model (Qwen3-14B-Base). As shown in Table \ref{tab:Scaling model size}, PretrainRL consistently enhances performance even for models with substantially larger parameter.  The increased parameter inherently strengthens the model’s learning capability and provides greater room for knowledge consolidation. This enables PretrainRL to more effectively unlock and refine the latent knowledge embedded in the LLM, particularly leveraging the enhanced representational power of larger architectures.

\begin{table}[h]
\small
\setlength{\tabcolsep}{3pt}
\centering
\caption{PretrainRL on EntityQuestions}
\vskip -0.1in
\resizebox{ \columnwidth}{!}{%
\begin{tabular}{l | c c c c}
\toprule
Method & \textbf{ACC (\%)} & \textbf{HR (\%)} & \textbf{MRR (\%)} & \textbf{Prob (\%)} \\
\midrule
Qwen3-4B & 16.39 & 44.53 & 20.49 & 7.88 \\
~~~~ {\em w/ PretrainRL} & 28.22 & 56.23 & 31.48 & 23.17 \\
\midrule
Qwen3-8B & 21.45 & 51.97 & 25.66 & 8.90 \\
~~~~ {\em w/ PretrainRL} & 31.39 & 61.87 & 35.02 & 25.46 \\
\bottomrule
\end{tabular}
}%
\label{tab:PretrainRL on EntityQuestions}
\vskip -0.1in
\end{table}

To further validate the effectiveness of our method on a larger-scale dataset, we evaluate it on EntityQuestions, a dataset sharing a long-tail distribution with POPQA but containing over 1.75 million questions. For each instance, we generate five negative example pairs, resulting in 8.7 million training instances for PretrainRL. As shown in Table \ref{tab:PretrainRL on EntityQuestions}, both models exhibit significant improvements on EntityQuestions, aligning with our results on other datasets. These consistency underscores the scalability and generalizability of our approach across varying model sizes and data scales.

\subsection{Ablation Study}
In this section, we conduct ablation studys to investigate the contributions of each component of PretrainRL to model performance. In our experiments, w/o NTP removes the next-token prediction (NTP) loss term from the total loss and computes only the DPO loss on the preferred pairs for preference optimization; w/o DPO removes the DPO loss and computes only the NTP loss, equivalent to standard continual pretraining. For fairness, all experiments use the same amount data. As reported in Table \ref{tab:Ablation Study}, removing the NTP loss (w/o NTP) reduces accuracy and hit rate but substantially increasing the probability of generating chosen (winning) responses. Meanwhile, removing the DPO loss (w/o DPO) helps consolidate the model’s factual knowledge, but in essence increases the frequency of long-tail knowledge by optimizing the corpus distribution. However, this approach risks suppressing other knowledge, as evidenced by the decreased Prob metric on the Wikidata-knowledge infusion dataset. In contrast, PretrainRL balances these trade-offs by learning knowledge during continual pretraining while reshaping the model’s probability distribution to make room for learning truthful responses, ultimately yielding comprehensive improvements.

\begin{table}[th]
\small
\setlength{\tabcolsep}{3pt}
\centering
\caption{Ablation Study}
\vskip -0.1in
\resizebox{ \columnwidth}{!}{%
\begin{tabular}{l | c c c c}
\toprule
Method & \textbf{ACC (\%)} & \textbf{HR (\%)} & \textbf{MRR (\%)} & \textbf{Prob (\%)} \\
\midrule
\multicolumn{5}{c}{\cellcolor{lightgray}\textit{PopQA}}\\
Qwen3-4B & 13.05 & 31.60 & 15.17 & 12.38 \\
~~~ {\em w/o NTP} & 9.67 & 18.06 & 11.02 & 33.88 \\
~~~ {\em w/o DPO} & 29.97& 60.36& 32.94 & 20.92 \\
~~~ {\em w/ PretrainRL} & 37.69 & 66.39 & 40.51 & 31.42 \\
\midrule
Qwen3-8B & 16.32 & 37.06 & 18.60 & 18.60 \\
~~~ {\em w/o NTP} & 13.08 & 22.17 & 13.92 & 42.11 \\
~~~ {\em w/o DPO} & 38.53 & 69.17 & 41.93 & 27.50 \\
~~~ {\em w/ PretrainRL} & 48.61 & 74.92 & 51.51 & 40.48 \\
\midrule
\multicolumn{5}{c}{\cellcolor{lightgray}\textit{wikidata}}\\
Qwen3-4B & 20.65 & 66.96 & 31.49 & 4.44 \\
~~~ {\em w/o NTP} & 33.84 & 47.46 & 37.54 & 18.42 \\
~~~ {\em w/o DPO} & 37.42& 79.32& 45.89 & 2.57 \\
~~~ {\em w/ PretrainRL} & 46.20 & 78.98 & 54.11 & 7.16 \\
\midrule
Qwen3-8B & 23.96 & 70.43 & 34.81 & 5.51 \\
~~~ {\em w/o NTP} & 38.51 & 49.49 & 41.70 & 51.41 \\
~~~ {\em w/o DPO} & 31.45 & 78.78 & 41.47 & 4.51 \\
~~~ {\em w/ PretrainRL} & 47.58 & 77.10 & 55.98 & 11.10 \\
\bottomrule
\end{tabular}
}%
\label{tab:Ablation Study}
\vskip -0.2in
\end{table}

We compare different sampling strategies to validate the effectiveness of our rejected sampling method. Concretely, we partition POPQA by quantiles of attribute-value popularity, designating the top 10\% of attribute values by popularity as \textit{head} attributes and treating the remainder as \textit{mid-tail} attributes. For each instance, rejected samples are randomly drawn from the head attribute pool. Aside from the sampling strategy, all experimental settings are kept identical. As shown in Apendix \ref{sec:Compare sampling methods}, our method demonstrates superior performance. Popularity-based sampling relies on an externally defined popularity metric and a rule-based partitioning of high-frequency knowledge, making it susceptible to dataset-specific distribution biases. In contrast, PretrainRL identifies high-frequency knowledge based on the model’s intrinsic probability distribution during generation, more faithfully reflecting the model’s internal knowledge.

\begin{table}[h]
\small
\setlength{\tabcolsep}{3pt}
\centering
\caption{Performance on different benchmarks}
\vskip -0.1in
\resizebox{ \columnwidth}{!}{%
\begin{tabular}{l|ccccc}
\toprule
Method & \textbf{CEval} & \textbf{MMLU} & \textbf{MATH} & \textbf{GSM8K}  & \textbf{BBH} \\
\midrule
Qwen3-4B & 50.81 & 52.39 & 14.46 & 83.62 & 71.80 \\
~~ {\em w/ PretrainRL} & 49.06 & 52.01 & 17.22 & 80.97 & 66.07 \\
~~ {\em w/ SFT} & 29.38 & 31.76 & 0.26 & 0.53 & 4.02 \\
\bottomrule
\end{tabular}
}%
\label{tab:Downstream work performance}
\vskip -0.2in
\end{table}

\subsection{Generalization Study}
\label{sec:general_study}
Prior studies \cite{charton2024emergentpropertiesrepeatedexamples,lv2025injectknowledgeefficientlyknowledge} have shown that excessive knowledge infusion does not always yield proportional benefits, and fine-tuning models for specific capabilities may inadvertently compromise performance on downstream tasks. To validate the generalizability and robustness of our method, we evaluate model trained with PretrainRL on diverse benchmarks: CEval \cite{huang2023ceval}, MMLU \cite{hendryckstest2021}, MATH \cite{wang-etal-2024-math}, GSM8K \cite{cobbe2021gsm8k}, and BBH \cite{suzgun2022challenging}. All evaluations are conducted using the OpenCompass \cite{2023opencompass}. As shown in Table \ref{tab:Downstream work performance}, PretrainRL does not cause any noticeable decline in overall performance, indicating that the improvements are not achieved at the expense of other abilities. In contrast, models fine-tuned via SFT exhibit substantial performance collapse on downstream tasks, highlighting PretrainRL’s unique ability to balance knowledge consolidation with general capability preservation.

\subsection{Case Study}
\begin{figure}[t]
\begin{center}
\centerline{\includegraphics[width=\columnwidth]{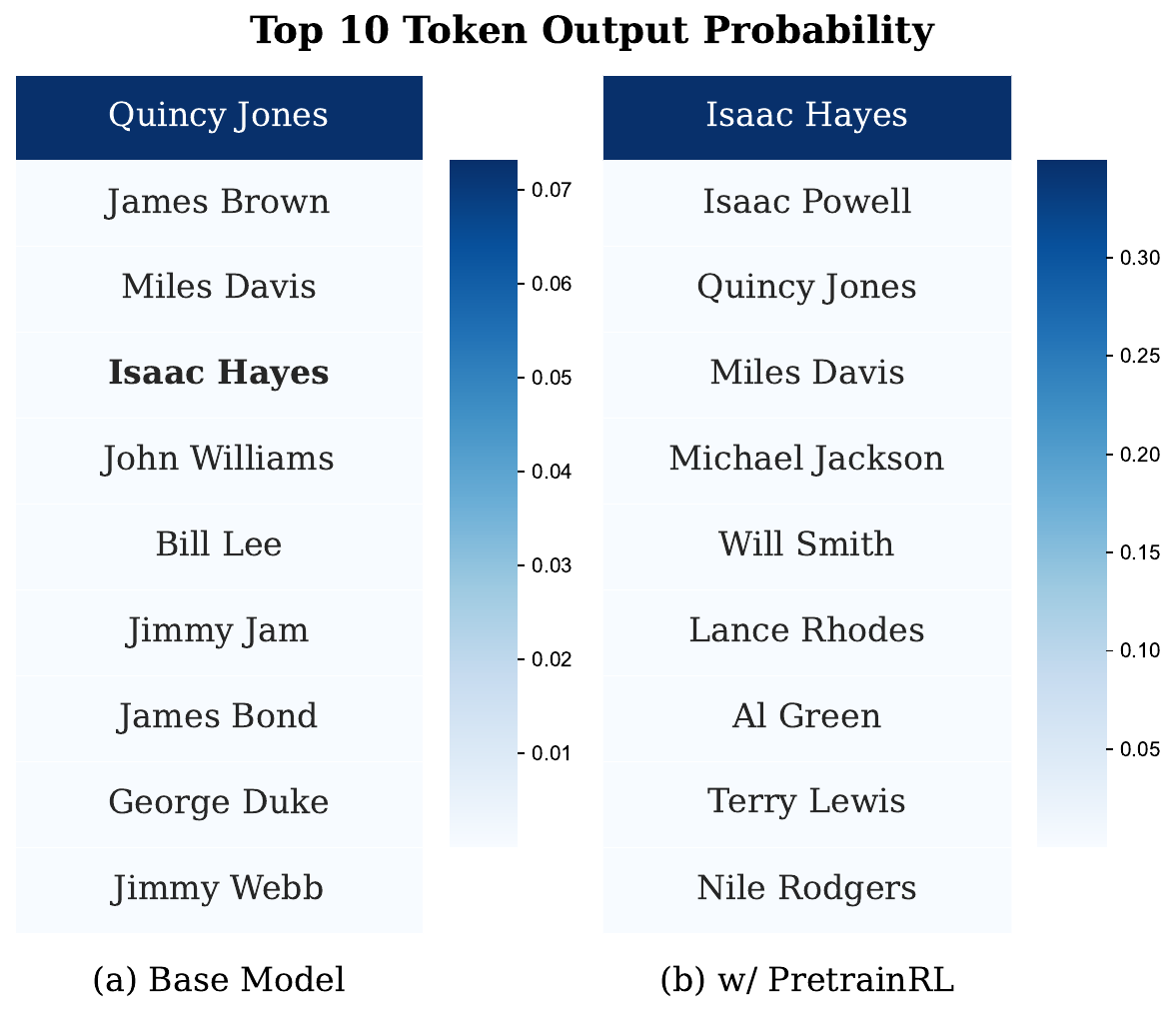}}
\caption{The output of the model for the question “Who was the composer of Shaft?”.}
\label{fig:Token_Probability}
\end{center}
\vskip -0.4in
\end{figure}

The imbalanced data distribution in training corpora leads LLM to exhibit a strong bias toward generating head knowledge when answering questions, while long-tail knowledge often degrades from conditional probability estimation to marginal probability approximation, resulting in severe hallucinations. In our method, we employ beam search decoding to obtain negative samples while simultaneously capturing the probability distribution of candidate answers. For instance, for the question \textit{“What is Yungay the capital of?”}, beam search retrieves the top 50 candidate answers,  including the ground truth \textit{"Yungay Province"}, which ranks 29th with a probability of $5.5802e-33$. This indicates that while the model retains latent memory of the correct answer, its marginal probability estimation fails to prioritize truthful responses. PretrainRL addresses this by debiasing the model’s probability distribution for long-tail knowledge, thereby mitigating hallucinations. Figure \ref{fig:Token_Probability} visualizes the probability distribution before and after training for a representative instance. At the beginning, the model initially assigns negligible probability to the correct answer. After PretrainRL, the answer’s probability is significantly recalibrated and stabilized, demonstrating the method’s effectiveness in aligning knowledge recall with factual accuracy.

\section{Related work}
Hallucination has been a fundamental challenge in LLMs and has received extensive attention in existing studies \cite{sahoo-etal-2024-comprehensive,kalai2025languagemodelshallucinate,liu2025eckgbench}. At the data level, prior research \cite{kamalloo2023hagridhumanllmcollaborativedataset,pal2023medhaltmedicaldomainhallucination} seek to reduce the interference caused by noise, factual inaccuracies, and conflicting data by constructing  higher-quality training corpora. Nevertheless, corpus imbalance is difficult to avoid. At the training strategy level, most work \cite{song2025hallucinationtaxreinforcementfinetuning} focuses on enabling models to say ``I don't know'' when they lack confidence in answering a question. Along this line, several approaches employ self-refinement mechanisms where models improve output accuracy through post-hoc feedback or internal reasoning processes. In addition, knowledge injection and Reinforcement Finetuning approaches have been explored to mitigate hallucinations. Notably, recent investigations \cite{gudibande2023falsepromiseimitatingproprietary,gekhman-etal-2024-fine} reveal that fine-tuning on unpopular factual knowledge may encourage hallucinations and LLMs struggle to integrate new knowledge through fine-tuning. To alleviate hallucinations concerning unpopular knowledge, we conduct an in-depth investigation from the perspective of the pretraining paradigm.

\section{Conclusion}
\label{sec:conclusion}

In this paper, we identify the imbalanced data distribution in pretraining corpora as a root cause of factual hallucinations.
To address this at its source, we propose PretrainRL, a novel framework that integrates reinforcement learning into the pretraining phase.
Guided by a ``debiasing then learning'' principle, PretrainRL leverages DPO and an efficient negative sampling strategy to reshape the model's probability distribution, making room for factual knowledge to be consolidated.
Extensive experiments on three public benchmarks demonstrate that PretrainRL significantly alleviates factual hallucinations.
Our work presents a paradigm shift towards a more fundamental ``pretrain-and-align'' approach for instilling factual reliability, moving beyond post-hoc corrections.

\section*{Limitations}
In our sampling procedure, to improve efficiency, we aggregate the head knowledge generated by the model for each question category and select the corresponding head knowledge as negative samples, which presupposes that the dataset offers a meaningful distinction among question categories. If the dataset lacks such differentiation,  the sampling strategy can be flexibly substituted with alternative methods such as RLVR to maintain robustness. This adaptability ensures that PretrainRL remains applicable across diverse data structures while preserving its core objective of mitigating hallucinations through targeted negative sampling.




\bibliography{custom}

\appendix
\clearpage
\section{Prompt}
\label{sec:Prompt}
\subsection{Question Answering Prompt (Few-shot)}
\begin{table}[ht!]
\centering
\scriptsize
\caption{Question answering (Few-shot).}
\vskip -0.05in
\begin{tabular}{p{0.45\textwidth}}
\toprule
Answer the following questions in as few words as possible.\\\\

Question: What is the capital of China?\\
Answer: Beijing\\\\

Question: What’s the job of Song Kang-ho in Parasite (2019)?\\
Answer: actor\\\\

Question: \{QUESTION\}\\
Answer: \\
\bottomrule
\end{tabular}
\label{table:Question Answering Prompt (Few-shot)}
\end{table}

\subsection{Question Answering Prompt (CoT)}
\begin{table}[ht!]
\centering
\scriptsize
\caption{Question answering (CoT).}
\vskip -0.05in
\begin{tabular}{p{0.45\textwidth}}
\toprule
Your task is to answer the question below. Give step by step reasoning before you answer,and when you’re ready to answer. Answer the question in as few words as possible.\\\\

Question: What is the capital of China?\\
Answer: Beijing\\\\

Question: What’s the job of Song Kang-ho in Parasite (2019)?\\
Answer: actor\\\\

Question: \{QUESTION\}\\
Answer: \\
\bottomrule
\end{tabular}
\label{table:Question Answering Prompt (CoT)}
\end{table}

\subsection{Training data template}

\begin{table}[ht!]
\centering
\scriptsize
\caption{Training data template}
\vskip -0.05in
\begin{tabular}{p{0.45\textwidth}}
\toprule
\{Question\}\{pos\}<pad>\{Question\}\{neg\}\\
\midrule
Example:\\
What is the capital of Arno? Florence<pad>What is the capital of Arno? Rome\\
\bottomrule
\end{tabular}
\label{table:prompt}
\end{table}

\section{Dataset}
\subsection{Dataset Description}
\label{sec:appendix Dataset Description}
POPQA\footnote{\url{https://github.com/AlexTMallen/adaptive-retrieval}} is created by sampling knowledge triples from Wikidata and converting them to natural language questions, followed by popularity calculation. The whole dataset consists of 14k questions. To avoid knowledge conflicts during the learning process, we deduplicated the POPQA dataset by retaining only subject entities that have a unique corresponding property entity, resulting in a final set of 13k questions. Wikidata-knowledge infusion\footnote{\url{https://huggingface.co/datasets/RJZ/wikidata_triple_en}} follows a similar construction process to POPQA, aggregating factual triplets from Wikidata and selecting six common relationship types to form 28k questions. To balance category disparities, we perform sampling on each category, yielding a final dataset of 16k questions. To validate the effectiveness of our method on a larger-scale, we also conduct experiments on EntityQuestions\footnote{\url{https://github.com/princeton-nlp/EntityQuestions}}, another popular open-domain QA dataset. Similar to POPQA, EntityQuestions uses Wikipedia hyperlink counts as a proxy for entity frequency and samples knowledge triples from Wikidata based on frequency distributions. After deduplicating subject entities, we obtain a final set of 1.75 million questions. The aforementioned datasets vary in size but all feature long-tail distributions. Following in POPQA, we analyze the popularity distribution of attribute values. 

\subsection{Statistics of the evaluation dataset}
\label{sec:Statistics of the evaluation dataset}
\begin{table}[th]
    \centering
    \setlength{\tabcolsep}{0.1cm}
    \caption{Statistics of POPQA}
    \vskip -0.05in
    \resizebox{1.0\columnwidth}{!}{
    \begin{tabular}{llc} 
    \toprule
    {Relation} & {Question template} & {Count} \\
    \midrule
    occupation & What is \{s\}’s occupation? & 531 \\
    father & Who is the father of \{s\}? & 562 \\
    country & In what country is \{s\}? & 824 \\
    director & Who was the director of \{s\}? & 1784 \\
    composer & Who was the composer of \{s\}? & 880 \\
    place of birth & In what city was \{s\} born? & 584 \\
    color & What color is \{s\}? & 34 \\
    religion & What is the religion of \{s\}? & 326 \\
    capital of & What is \{s\} the capital of? & 255 \\
    producer & Who was the producer of \{s\}? & 1320 \\
    author & Who is the author of \{s\}? & 1408 \\
    genre & What genre is \{s\}? & 1530 \\
    screenwriter & Who was the screenwriter for \{s\}? & 1671 \\
    mother & Who is the mother of \{s\}? & 187 \\
    capital & What is the capital of \{s\}? & 622 \\
    \midrule
    \multicolumn{2}{l}{Total-Count} & 12,518 \\
    \bottomrule
    \end{tabular}
    }
    \label{tab:Statistics of the evaluation dataset}
\end{table}

\begin{table}[th]
    \centering
    \setlength{\tabcolsep}{0.1cm}
    \caption{Statistics of Wikidata-knowledge infusion}
    \vskip -0.05in
    \resizebox{1\columnwidth}{!}{
    \begin{tabular}{llc} 
    \toprule
    {Relation} & {Question template} & {Count} \\
    \midrule
    capital & What is the capital of \{s\}? & 421 \\
    industry & What is the industry of \{s\}? & 4000 \\
    location & What is the location of \{s\}? & 4000 \\
    material & What is the material of \{s\}? & 4000 \\
    color & What is the color of \{s\}? & 3627 \\
    shape & What is the shape of \{s\}? & 342 \\
    \midrule
    \multicolumn{2}{l}{Total-Count} & 16,390 \\
    \bottomrule
    \end{tabular}
    }
    \vskip 0.2in
    \label{tab:Statistics of Wikidata}
\end{table}

\begin{table}[th]
    \centering
    \setlength{\tabcolsep}{0.1cm}
    \caption{Statistics of EntityQuestions}
    \vskip -0.05in
    \resizebox{1.0\columnwidth}{!}{
    \begin{tabular}{llc} 
    \toprule
    {Relation} & {Question template} & {Count} \\
    \midrule
    P36 & What is the capital of \{s\}? & 6704 \\
    P407 & Which language was \{s\} written in? & 5113 \\
    P26 & Who is \{s\} married to? & 7979 \\
    P159 & Where is the headquarter of \{s\}? & 7994 \\
    P276 & Where is \{s\} located? & 7902 \\
    P40 & Who is \{s\}'s child? & 7981 \\
    P176 & Which company is \{s\} produced by? & 7913 \\
    P20 & Where did \{s\} die? & 7985 \\
    P112 & Who founded \{s\}? & 4060 \\
    P127 & Who owns \{s\}? & 7936 \\
    P19 & Where was \{s\} born? & 7991 \\
    P740 & Where was \{s\} founded? & 7506 \\
    P413 & What is \{s\} famous for? & 7971 \\
    P800 & What position does \{s\} play? & 1762 \\
    P69 & Where was \{s\} educated? & 7990 \\
    P50 & Who is the author of \{s\}? & 7945 \\
    P170 & Who was \{s\} created by? & 6709 \\
    P106 & What kind of work does \{s\} do? & 7988 \\
    P131 & Where is \{s\} located? & 7921 \\
    P17 & Which country is \{s\} located in? & 7976 \\
    P175 & Who performed \{s\}? & 7767 \\
    P136 & What type of music does \{s\} play? & 7954 \\
    P264 & What music label is \{s\} represented by? & 7841 \\
    P495 & Which country was \{s\} created in? & 7909 \\
    \midrule
    \multicolumn{2}{l}{Total-Count} & 174,797 \\
    \bottomrule
    \end{tabular}
    }
    \label{tab:Statistics of EntityQuestions}
\end{table}

\section{Compare different sampling methods}
\label{sec:Compare sampling methods}
\begin{table}[th]
\small
\setlength{\tabcolsep}{3pt}
\centering
\caption{Compare sampling methods on POPQA}
\vskip -0.1in
\resizebox{ \columnwidth}{!}{%
\begin{tabular}{l | c c c c}
\toprule
Method & \textbf{ACC (\%)} & \textbf{HR (\%)} & \textbf{MRR (\%)} & \textbf{Prob (\%)} \\
\midrule
Qwen3-4B & 16.39 & 44.53 & 20.49 & 7.88 \\
Popularity-based sampling & 24.75 & 48.98 & 26.93 & 27.72 \\
PretrainRL(ours) & 37.69 & 66.39 & 40.51 & 31.42 \\
\midrule
Qwen3-8B & 21.45 & 51.97 & 25.66 & 8.90 \\
Popularity-based sampling & 35.47 & 60.97 & 37.93 & 34.95 \\
PretrainRL(ours) & 48.61 & 74.92 & 51.51 & 40.48 \\
\bottomrule
\end{tabular}
}%
\label{tab:compare sampling methods on popQA}
\vskip -0.1in
\end{table}

\end{document}